\documentclass[11pt,a4paper]{article}
\usepackage[hyperref]{eacl2021}
\usepackage{times}
\usepackage{latexsym}
\usepackage{graphicx}
\graphicspath{ {images/} }
\usepackage{CJKutf8}
\usepackage{amsmath}
\usepackage{amssymb}
\usepackage{bm}

\newcommand{\argmax}{\mathop{\rm arg~max}\limits}

\usepackage{microtype}

\aclfinalcopy 

\title{Hierarchical Multitask Learning with Dependency Parsing for Japanese Semantic Role Labeling Improves Performance of Argument Identification}

\author{Tomohiro Nakamura, Tomoya Miyashita, and Soh Ohara \\
  pluszero Inc., Tokyo, Japan \\
  The University of Tokyo, Japan\\
  \texttt{nakamura.tomohiro@plus-zero.co.jp}, \texttt{miyashita.tomoya@plus-zero.co.jp},\\
  \texttt{ohara.soh@plus-zero.co.jp}}

\date{}

\begin{document}
\maketitle

\begin{abstract}
With the advent of FrameNet and PropBank, many semantic role labeling (SRL) systems have been proposed in English. 
Although research on Japanese predicate argument structure analysis (PASA) has been conducted, most studies focused on surface cases. 
There are only few previous works on Japanese SRL for deep cases, and their models' accuracies are low.
Therefore, we propose a hierarchical multitask learning method with dependency parsing (DP) and show that our model achieves state-of-the-art results in Japanese SRL.
Also, we conduct experiments with a joint model that performs both argument identification and argument classification simultaneously. The result suggests that multitasking with DP is mainly effective for argument identification.
\end{abstract}

\section{Introduction}
\begin{figure}
  \includegraphics[width=7.5cm]{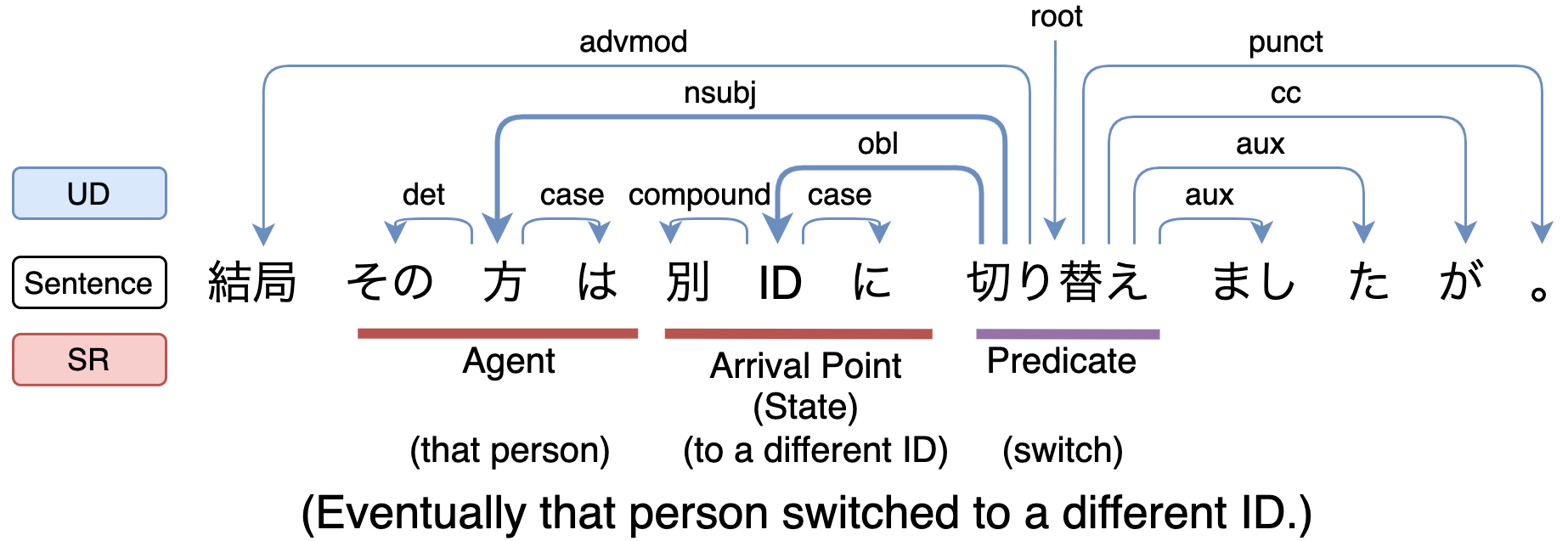}
  \caption{An example sentence annotated with UD (blue) and  semantic roles (red).}
  \label{fig:srl-and-ud}
\end{figure}

\begin{CJK}{UTF8}{min}
Semantic role labeling (SRL) is a kind of predicate argument structure analysis (PASA), which is a task to identify predicates and their corresponding arguments in a sentence and assign an appropriate semantic tag (a semantic role) to each argument.
SRL can be divided into three sub-tasks: predicate detection, argument identification, and argument classification\footnote{Some papers include predicate sense disambiguation, but we exclude it because BCCWJ-PT does not adopt predicate labels.}.
Predicate detection detects a predicate span, argument identification detects argument spans for the predicate detected, and argument classification assigns semantic roles to the arguments detected.
For example, in Figure \ref{fig:srl-and-ud}, we first identify  "切り替え"  (switch) as a predicate. Then we identify "その方は" (that person) and "別IDに" (to a different ID)  as the arguments and assign the semantic roles \textit{Agent} and \textit{Arrival Point (State)} to them, respectively.
Analyzing the semantic relationship between predicates and arguments is essential in various natural language processing applications, such as machine reading comprehension \citep{zhang2019semantics, guo-etal-2020-frame}.

\end{CJK}

\begin{CJK}{UTF8}{min}
Predicate and argument relation labels can be divided into two main categories: surface cases, using case markers in sentences, and deep cases, taking into account more semantic aspects.
In English, SRL shared tasks \citep{litkowski-2004-senseval, carreras-marquez-2005-introduction, hajic-etal-2009-conll, pradhan-etal-2012-conll} have been held for PropBank \citep{kingsbury2002treebank} and FrameNet \citep{baker-etal-1998-berkeley-framenet}, which adopt deep cases.
However, in Japanese, almost all studies
\citep{ouchi-etal-2015-joint, shibata-etal-2016-neural, ouchi-etal-2017-neural,matsubayashi-inui-2017-revisiting, matsubayashi-inui-2018-distance, omori-komachi-2019-multi} 
have been focused on only three surface cases: nominative case (が; ga), accusative case (を; wo), and dative case (に; ni).
Some studies \citep{okamura-etal-2018-improving, weko_200684_1} tackled Japanese SRL for deep cases, including semantic roles such as time, factor, and location, using BCCWJ-PT \citep{takeuchi2015annotating}. 
\citet{okamura-etal-2018-improving} proposed neural network models by applying transfer learning using a different SRL corpus (GDA corpus\footnote{https://www.gsk.or.jp/catalog/gsk2009-b/}).
However, they assumed argument spans are given. Hence their studies are limited to argument classification only.
In most NLP tasks, argument spans are not given in advance. Thus the models that do not perform argument identification are not practical.
Therefore, this paper proposes a model that performs argument identification and argument classification jointly by using BIO tags in Japanese. 
\end{CJK}

Furthermore, we focus on the relationship between SRL and universal dependency (UD) representations.
\begin{CJK}{UTF8}{min}
For example, in Figure \ref{fig:srl-and-ud}, the edges $\langle$"切り替え" (switch), "方" (person)$\rangle$ and $\langle$"切り替え" (switch), "ID"$\rangle$ correspond to \textit{Agent} and \textit{Arrival Point (State)}, respectively.
We can see that the dependency trees can provide essential information for argument identification, and the types of edges are beneficial for argument classification.
In this paper, we utilize the information of the UD dataset implicitly through multitasking.
Also, the usage of the UD dataset is preferable due to the deficiency of Japanese SRL data (Japanese UD treebank is around 10 times of BCCWJ-PT).

\end{CJK}

This paper's contributions are the following:
(1) we propose the first hierarchical multitask model that combines Japanese DP and SRL, and it achieves state-of-the-art results in Japanese SRL;
(2) by assigning BIO tags to each morpheme, our model performs argument identification and argument classification simultaneously in Japanese SRL;
(3) we show that multitasking with DP improves the performance of  argument identification.

\begin{figure}
  \includegraphics[width=6cm]{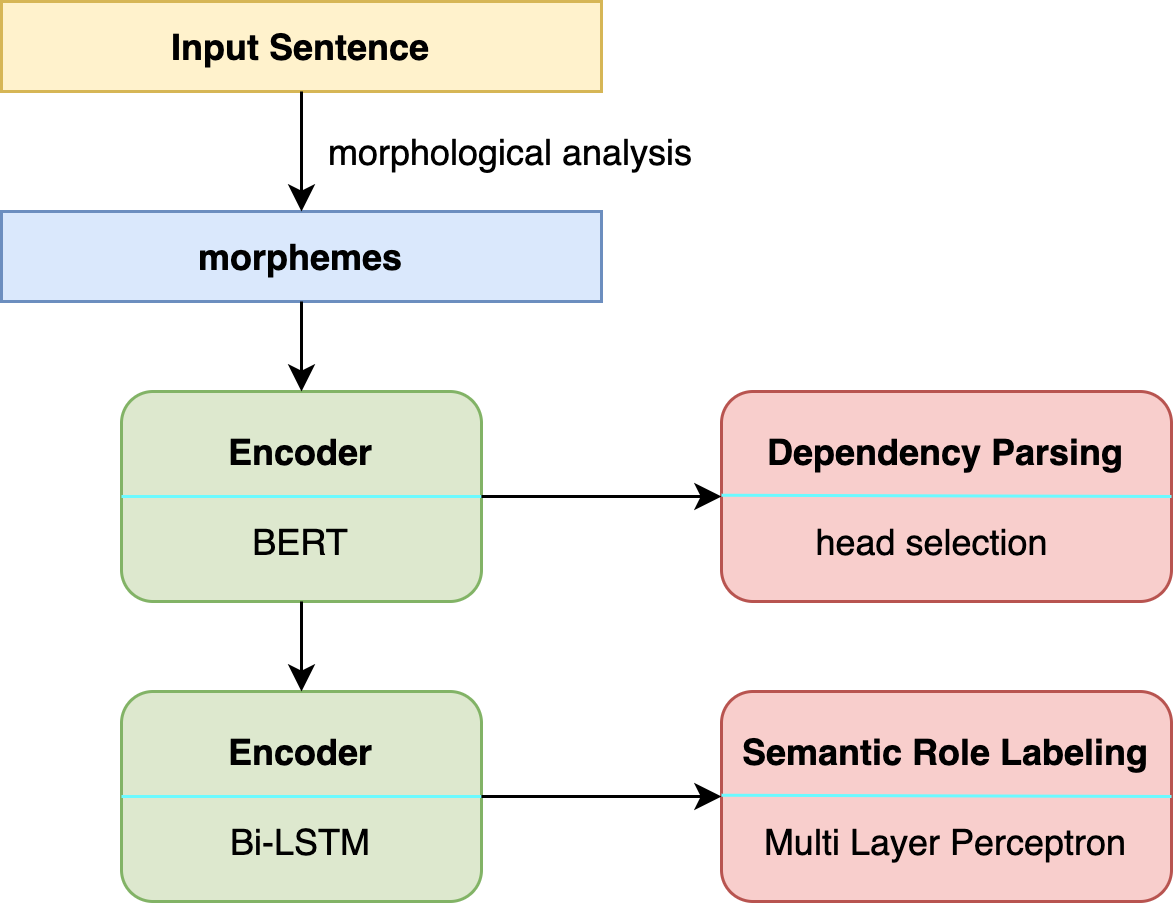}
  \caption{Diagram of the model architecture.}
  \label{fig:model-arc}
\end{figure}

\section{Model}
In this section, we describe our hierarchical multitask model. 
Figure \ref{fig:model-arc} shows an overview of the hierarchical model.
BERT \citep{devlin2018bert} is used as a shared encoder for DP and SRL, and BiLSTM is used for SRL only.

\subsection{Japanese Morphemes}
Unlike English, there are no spaces between words in Japanese.
Therefore, to transform a sentence into input vectors, it is necessary to separate it into morphemes. In this paper, UniDic Short Unit Word (SUW) morphemes\footnote{https://unidic.ninjal.ac.jp/} are converted into subword tokens with Byte Pair Encoding (BPE) \cite{sennrich-etal-2016-neural}, and they are used as inputs to the models.
The output vectors of the models are averaged over the specified units for each task.

When training and evaluating the DP model, we assume sentences are tokenized, and the true SUW morphemes are given.
In contrast, when training and evaluating the SRL model, we tokenize sentences by MeCab\footnote{https://taku910.github.io/mecab/} \citep{kudo-etal-2004-applying} with unidic-cwj and split them into SUW morphemes.
This is because true Unidic Long Unit Word (LUW) morphemes are given in SRL dataset, but the true SUW morphemes are not.

We adopt SUW as the basic morphological unit for DP, and we design the \textit{morpheme setting} and the \textit{span given setting} for SRL.
In the \textit{morpheme setting}, our model performs both argument identification and argument classification, giving BIO-tagged semantic roles to all LUW morphemes in the sentence. In the \textit{span given setting}, argument spans are already given, and our models only perform argument classification. 

\subsection{Dependency Parsing}
We consider dependency parsing (DP) as a head selection problem \citep{zhang-etal-2017-dependency-parsing, dep-shibata-parsing}.
Thus, our model predicts the most likely head for each token.
Although \citet{zhang-etal-2017-dependency-parsing} built a model that only performs label classification, our model performs head selection and label prediction by multitasking.

We define the \textit{root unknown setting} and the \textit{root known setting} for the model's input.
In the \textit{root unknown setting}, the input is $S_{dp} = \{[CLS], w_1, ..., w_n, [SEP], [ROOT]\}$, where $\{w_1, ..., w_n\}$ are SUW morphemes in the sentence, and $[ROOT]$ is a special token representing the root.
In the \textit{root known setting}, the input is $S_{multi\_dp} = \{[CLS], w_1, ..., w_n,[SEP],$ $w_{root}, [SEP],[ROOT]\}.$
This input mimics the input of SRL (described in \ref{seq:srl}) to maximize the effect of multitasking. 
However, the \textit{root known setting} explicitly takes $w_{root}$ as the input while DP should be solved without knowing the root location.
Therefore, in the \textit{root known setting}, we use the true $w_{root}$ during training, whereas we use the $w_{root}$ predicted by a DP single-task model trained in the \textit{root unknown setting} during testing.

\subsection{Semantic Role Labeling}
\label{seq:srl}
In this paper, the predicate is assumed to be already given, so we only discuss argument identification and argument classification.
Our SRL model structure is the same as \citet{shi2019simple} which is composed of BERT and BiLSTM encoder. We simultaneously performed argument identification and argument classification in Japanese as assigning BIO-tagged semantic role labels (e.g., \textit{B-Agent}, \textit{I-Agent}, \textit{O}) to SUW morphemes, following
\citet{shi2019simple}, who gave BIO tags to words in English.

\begin{CJK}{UTF8}{min}
When performing SRL, BERT takes $S_{srl} = \{[CLS], w_1, ..., $ $w_n,[SEP], w_{p}, [SEP]\}$ as the input.
The output hidden vectors of BERT are concatenated with the predicate indicators, which indicate whether each token is a predicate or not. The combined vectors are passed to a one-layer BiLSTM, and the hidden vectors of the ouput are averaged over the suitable units for SRL (a LUW morpheme in the \textit{morpheme setting} and an argument span in the \textit{span given setting}).
The hidden vector of the predicate in the sentence is concatenated to each hidden vector, and it is fed into a one-hidden-layer MLP classifier over the label set.
\end{CJK}

\subsection{Loss and Training}
We use cross entropy loss for DP ($J_{dp}$) and SRL ($J_{srl}$).
The loss for DP is the sum of the loss of head selection and label prediction.
We use scaling factor $\lambda_{dp}$ and train the multitask models with $\lambda_{dp}J_{dp}$ and $J_{srl}$.

We follow the training method described in \citep{sogaard2016deep, subramanian2018learning, sanh2019hierarchical}:
after each parameter is updated, we choose one task and extract one batch of data to optimize the model.
We choose DP and SRL tasks with a ratio of $1-\beta_{srl}: \beta_{srl}$.

We use Optuna \citep{optuna_2019} to search for the best combination of hyperparameters automatically.
When we train the DP model, we set UAS as the target score. When we train the SRL or DP+SRL model, we set micro F1 scores of SRL.

\section{Experimental Setup}
We use the NICT BERT Japanese Pre-trained Model\footnote{https://alaginrc.nict.go.jp/nict-bert/index.html}, a bert-base model pre-trained on Japanese Wikipedia.

To evaluate our models' performance, we use Universal Dependencies (UD) Japanese treebank \citep{asahara-etal-2018-universal} for DP and BCCWJ-PT for SRL.
To compare our model with the baselines \citep{okamura-etal-2018-improving, weko_200684_1}, we split the BCCWJ-PT data with the same ratio 65:5:30 (train:dev:test) as them. Their models are only evaluated in the \textit{span given setting}, thus we only compare models in this setting.

\section{Result and Discussion}

\begin{table*}
    \centering
    \small
    \begin{tabular}{lllllll}
    \hline
    Model & UAS & LAS & micro $F1_{spa}$ & macro $F1_{spa}$ & micro $F1_{mor}$ & macro $F1_{mor}$\\
    \hline \hline
    \textbf{DP root unknown} & \bm{$94.84$} & $93.23$  &  - & - & - & - \\
    \textbf{DP root known} & $94.82$ & \bm{$93.29$}  &  - & - & - & - \\
    \textbf{SRL span given} & - & - & $75.98$ & 58.74 & - & -  \\
    \textbf{SRL  morpheme} & - & - & - & - & $58.55$ & $46.04$ \\
    \textbf{DP$+$SRL span given} & $94.54$ & $92.92$ & \bm{$77.48$} & \bm{$60.68$} & - & -  \\
    \textbf{DP$+$SRL morpheme} & $89.04$ & $85.94$ & - & - & \bm{$60.12$} & \bm{$47.77$} \\
    \hline
    \end{tabular}
    \caption{\label{citation-guide}
    The scores of each model. Models are trained and tested five times.
    $F1_{spa}$ and $F1_{mor}$ represent SRL F1 scores in the \textit{span given setting} and the \textit{morpheme setting}, respectively.
    }
    \label{tab:main_result}
\end{table*}

\begin{table}[]
    \centering
    \small
    \begin{tabular}{ll}
    \hline 
    Model & accuracy \\
    \hline \hline
    \citet{okamura-etal-2018-improving} & 66.5\\
    \citet{weko_200684_1} & 70.2 \\
    \textbf{SRL span given (65:5:30)} & \bm{$75.19$} \\
    \hline
    \end{tabular}
    \caption{The accuracy scores for the baseline setting. \citet{okamura-etal-2018-improving} is the CNN model, and \citet{weko_200684_1} is the Bi-GRU model, which are used as the baseline models, and we cited their scores from each paper.
    \textbf{SRL span given (65:5:30)} is the result used for a comparison with the baselines. Note that the only difference between \textbf{SRL span given (65:5:30)} and \textbf{SRL span given} is how we divided the SRL data (BCCWJ-PT).}
    \label{tab:baselines}
\end{table}

The main results are shown in Table \ref{tab:main_result}. \textbf{DP+SRL morpheme} and \textbf{DP+SRL span given} models are the multitask models, and other models are single-task models.
Multitask models' input is \textit{root known setting}.
Table \ref{tab:baselines} shows the result of our SRL single-task model and baselines.

\subsection{Result of Dependency Parsing}
UAS and LAS of DP single-task models (\textbf{DP root unknown} and \textbf{DP root known} in Table \ref{tab:main_result}) were almost the same. 
However, multitasking with SRL did not improve LAS and UAS of \textbf{DP root known} model. Especially in the \textit{morpheme setting}, multitasking instead worsened the score. 
This is because the loss scale for DP chosen by Optuna was small in the \textit{morpheme setting} ($\lambda_{dp} = 1.07 \times 10^{-2}$), and models were not trained enough to achieve the highest UAS and LAS scores.

\subsection{Results of Semantic Role Labeling}
Comparing with the baselines, our SRL model (\textbf{SRL span given (65:5:30)}) achieved a better accuracy of +4.99 in the \textit{span given setting} (Table \ref{tab:baselines}). Furthermore, we improved \textbf{SRL span given} model using multitasking with DP by +1.50 and +1.94 in micro and macro F1, respectively (Table \ref{tab:main_result}).

In the \textit{morpheme setting}, micro and macro F1 scores of the multitask model (\textbf{DP$+$SRL morpheme}) were higher than the single-task model by +1.57 and +1.73, respectively (Table \ref{tab:main_result}). 

\subsection{Contribution of Multitasking}
\begin{figure}
    \centering
    \includegraphics[width=7.5cm]{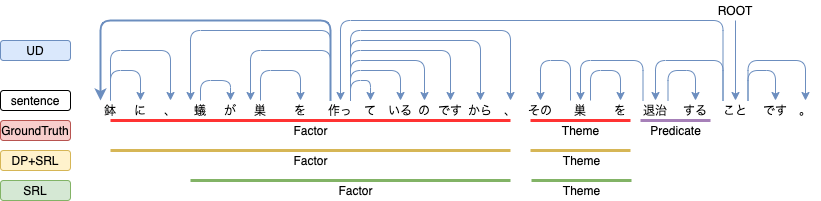}
    \caption{The specific result for test data of "鉢に、蟻が巣を作っているのですから、その巣を退治することです。" (In the pot, the ants have built a nest, and you have to remove the nest.).}
    \label{fig:case_study}
\end{figure}

\begin{table}
    \centering
    \small
    \begin{tabular}{lll}
        \hline
        setting & identification &  classification\\
        \hline \hline
        \textbf{SRL morpheme} & $72.87$ & $80.35$  \\
        \textbf{DP+SRL morpheme} & $\bm{74.46}$ & $\bm{80.74}$ \\
        \hline
    \end{tabular}
    \caption{The score for each step in SRL using morpheme unit. Argument identification score is F1 score, and argument classification score is accuracy.}
    \label{tab:srl_morpheme_detail_score}
\end{table}
\begin{CJK}{UTF8}{min}
Table \ref{tab:srl_morpheme_detail_score} shows the scores for argument identification and argument classification in the \textit{morpheme setting}. Accuracy of argument classification was calculated only using the samples that are correctly argument-identified. 
Using the unpaired t-test, there was a significant difference between single-task and multitask model in scores for argument identification (p-value=0.0008), but no significant difference in scores for argument classification (p-value=0.423).
This indicates that multitask learning with DP improved the score for argument identification. 
There were cases where the DP+SRL multitask model answered correctly, while the SRL single-task model made an error at the argument identification step.
One example is shown in Figure \ref{fig:case_study}. We can observe that "鉢に、" (in the pot,) is connected to "作っ" (make). This suggests that such a relationship between DP and SRL might be beneficial for the DP+SRL model to carry out argument identification.

\end{CJK}

\section{Related Work}
In Japanese, Kyoto Corpus \citep{kawahara2002construction} and NAIST Text Corpus \citep{iida2007annotating} have been widely used to evaluate the performance of PASA models for three surface cases. In contrast, EDR corpus\footnote{http://www2.nict.go.jp/ipp/EDR/ENG/indexTop.html}, BCCWJ-PT, Japanese FrameNet \citep{ohara2003japanese}, and GDA corpus are corpora with a variety of semantic roles. We used BCCWJ-PT to evaluate our models' performance.

In Japanese, PASA models did not perform "argument identification" in the same way we have defined.
In the papers \citep{identification-2010, ouchi-etal-2015-joint}, they mentioned argument identification, but the argument spans are already given, and they only selected a span corresponding to each superficial case.
To the best of our knowledge, our paper is the first work to solve such limitation and specify appropriate argument spans from sentences in Japanese by using neural network.



Syntactic information is important in SRL, and many papers have shown that passing syntactic information can improve SRL scores \citep{roth2016neural,marcheggiani-titov-2017-encoding, he-etal-2018-syntax}.
Also, it was reported that multitask learning with DP improves the accuracy of SRL \citep{peng-etal-2017-deep, cai2019syntax}. We implemented a hierarchical multitask model for UD and SRL similar to that for semantic tasks \citep{sanh2019hierarchical}.


\section{Conclusion}
We proposed a hierarchical multitask model that performs DP and SRL. Our model achieved state-of-the-art results in Japanese SRL for deep cases. Besides, our model performed argument identification and argument classification jointly by performing SRL at the morphological unit. Based on the empirical results, we show that multitasking is effective for argument identification.

\section*{Acknowledgments}
We would like to thank Y. Miyao and K. Hanaki for useful discussions.
We are grateful to A. Notoya, Y. Lee, K. Fujita, R. Mori, and M. Nagata for checking the contents of our paper and giving us appropriate advice.

\bibliography{anthology,eacl2021}
\bibliographystyle{acl_natbib}

\appendix

\section{Datasplit}

Distribution of UD Japanese treebank and BCCWJ-PT are shown in Table \ref{tab:ud-split} and Table \ref{tab:split-srl}. 
Both datasets are based on Balanced Corpus of Contemporary Written Japanese (BCCWJ) \citep{maekawa2014balanced}.
Thus, sentences in one's training data may appear in another's test data, which results in undesirable information leak.
To prevent such possible circumstances, BCCWJ-PT data shared with the DP training, validation, and test data are added to the SRL training, validation, and test data, respectively. The dataset is divided with a ratio of 80:10:10 as showin in Table \ref{tab:split-srl}.

\begin{table}
\centering
    \begin{tabular}{llll} \hline
       & training & validation & test \\ \hline \hline
      sentence & 40801 & 8427 & 7811\\ \hline
      morpheme & 923761 & 180767 & 168759\\ \hline 
    \end{tabular}
    \caption{Data distribution of UD Japanese treebank.}
    \label{tab:ud-split}
\end{table}


\begin{table}
\centering
    \begin{tabular}{llll} \hline
      & training & validation & test \\ \hline \hline
      total & 4055 & 507 & 507\\ \hline
      shared & 1784 & 477 & 486\\ \hline
    \end{tabular}
    \caption{The number of sentences and the number of shared sentences with the UD Japanese treebank in BCCWJ-PT.}
    \label{tab:split-srl}
\end{table}

\section{Hyperparameters}
The hyperparameters are the learning rate ($\eta$), the dropout probabilities ($\gamma_{bert}$, $\gamma_{dp}$, and $\gamma_{lstm}$), the scale of DP loss ($\lambda_{dp}$), and the probability that SRL batch is sampled while training ($\beta_{srl})$. 
As shown in Table \ref{tab:hparam}, the hyper parameters are tuned via Optuna, and their optimal values are shown in Table \ref{tab:optimized_params}.
Optuna is an optimization software available under MIT license\footnote{https://github.com/optuna/optuna}. 
We run the program for 50 times to maximize the target score of the validation data. 

\begin{table}
\centering
    \begin{tabular}{ c||c } \hline
      name & search space   \\ \hline \hline
      $\eta$ & logarithmic \\ \hline
      $\gamma_{bert}$, $\gamma_{dp}$, $\gamma_{lstm}$ & linear \\ \hline
      $\lambda_{dp}$ & logarithmic  \\
      \hline 
      $\beta_{srl}$ & linear\\ \hline
    \end{tabular}
    \caption{Hyperparameters tuned via optuna. $\eta$ is the learning rate. $\gamma_{bert}$, $\gamma_{dp}$, $\gamma_{lstm}$ are the dropout ratios. $\lambda_{dp}$ is the scale of DP loss. $\beta_{srl}$ is the proportion of SRL sample during training.}
    \label{tab:hparam}
\end{table}

\begin{table*}[]
    \centering
    \small
    \begin{tabular}{lllllll}
        \hline
        model & $\eta$ & $\gamma_{bert}$ & $\gamma_{dp}$ & $\gamma_{lstm}$ & $\lambda_{dp}$ & $\beta_{srl}$ \\
        \hline \hline
        \textbf{DP+SRL morpheme} & $9.34\times 10^{-5}$ & $0.130$ & $0.497$ & $0.431$ & $1.07\times 10^{-2}$ & $0.720$\\ \hline
        \textbf{DP+SRL span given} & $6.90\times 10^{-5}$ & $2.01\times 10^{-2}$ & $0.414$ & $0.202$ & $1.10$ & $0.314$\\ \hline
        \textbf{DP root unknown} & $7.77 \times 10^{-5}$ &  $0.111$ & $1.29 \times 10^{-3}$ & $-$ & $9.58 \times 10^{-2}$ & $-$\\ \hline
        \textbf{DP root known} & $7.90 \times 10^{-5}$ & $3.50\times 10^{-2}$ & $4.36 \times 10^{-5}$ & $-$ & $0.188$ &  $-$ \\ \hline
        \textbf{SRL morpheme} & $9.57 \times 10^{-5}$ & $1.99 \times 10^{-5}$ & $-$ &  $0.347$ & $-$ & $-$ \\ \hline
        \textbf{SRL span given} & $9.57\times 10^{-5}$ & $1.99 \times 10^{-5}$ & $-$ & $0.347$ & $-$ & $-$ \\ \hline
        \textbf{SRL span given(65:5:30)} & $8.41 \times 10^{-5}$ & $7.61\times 10^{-2}$ & $-$ & $0.667$ & $-$ & $-$ \\ \hline 
        \textbf{SRL morpheme} & $9.99\times 10^{-5}$ & $4.26\times 10^{-2}$ & $-$ & $6.79\times 10^{-3}$ & $-$ & $-$ \\ \hline 
    \end{tabular}
    \caption{Optimized hyperparameters for each models.}
    \label{tab:optimized_params}
\end{table*}

\section{Detailed Model}
\begin{CJK}{UTF8}{min}
\begin{figure*}
\centering
  \includegraphics[width=10cm]{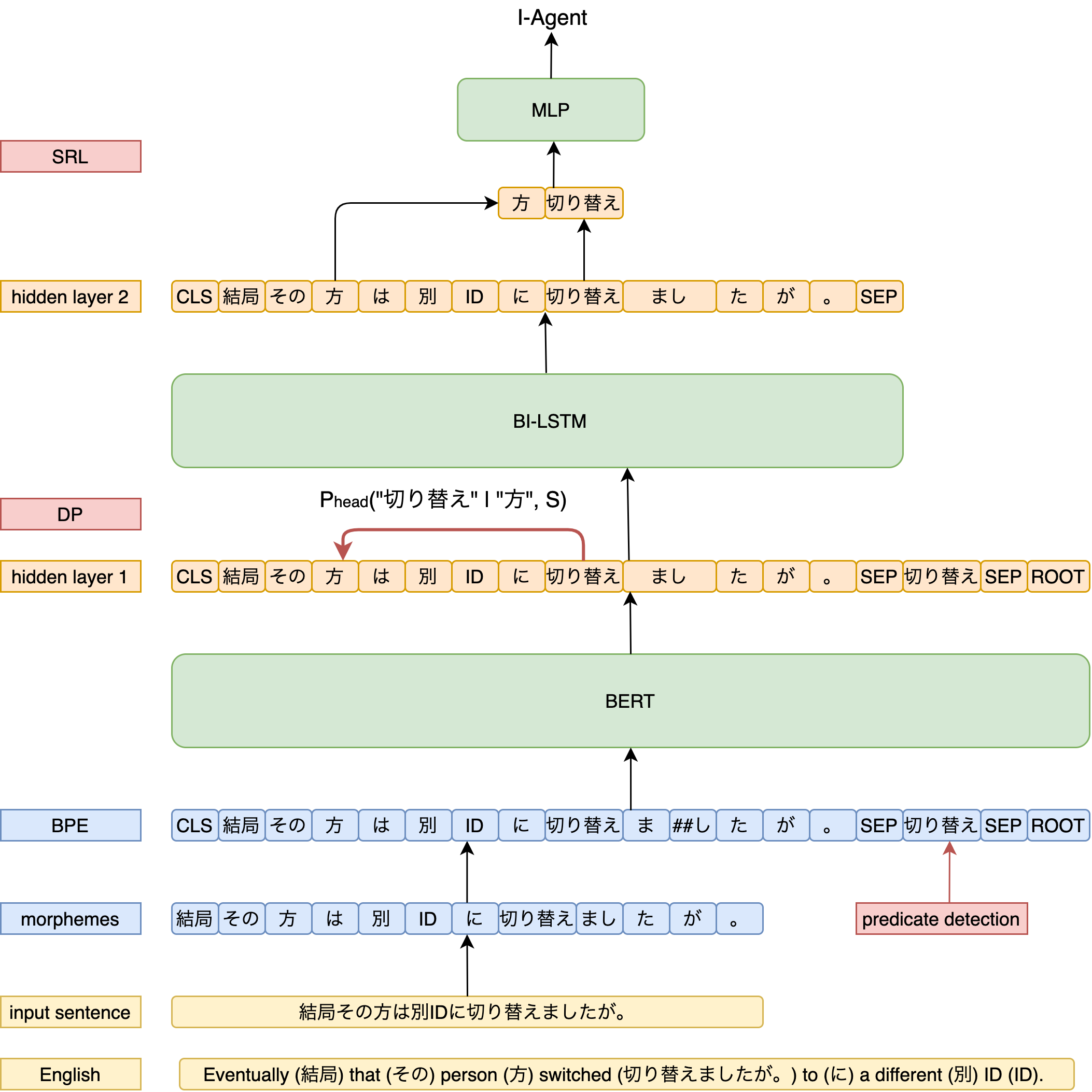}
  \caption{Architecture of our hierarchical multitask model. The model is making a prediction for the morphemes "方" at the final layer. }
  \label{fig:ud-model}
\end{figure*}
\end{CJK}

Figure \ref{fig:ud-model} shows the architecture of our hierarchical multitask model.


\subsection{Dependency Parsing}

Dependency parsing (DP) is a task that takes a sentence of length N as input and produces N $\langle$ head, dependent $\rangle$\  edges.
In a dependency tree, each head can have multiple dependents, while each dependent can only have one head.
Based on this relationship, we consider DP as a head selection problem.
We define morphemes of a sentence as $S=\{w_1, ..., w_n\}$, and let $S_{root}$ be $S_{root}=\{w_1, ..., w_n, [ROOT]\}$. We aim to choose the most appropriate head $w_j\in S_{root}$ for each dependent
$w_i\in S$. 
In $S_{root}$, [ROOT] does not have a head, but it is possible to choose [ROOT] as a head.
Let 
$X_{dp} = \{\bm{x}_1, ..., \bm{x}_n, \bm{x}_{[ROOT]}\}$
be the output of BERT for $S_{root}$
($X_{dp}$ is a hidden layer 1 in Figure \ref{fig:ud-model}),  then the probability that $w_j$ is the head of $w_i$ is
\begin{equation}
    P_{head}(w_j|w_i, S) = \frac{\exp(s(\bm{x}_j, \bm{x}_i))}{\sum_{k} \exp(s(\bm{x}_k, \bm{x}_i))}.
\end{equation}
The score $s$ is a function using a one-layer neural network and is calculated as
\begin{equation}
s(\bm{x}_j, \bm{x}_i) = \bm{v}^{T}\tanh(\bm{U} \bm{x}_j + \bm{W} \bm{x}_i),
\end{equation}
where $\bm{x}_i\in\mathbb{R}^{h}$ and $\bm{x}_j\in\mathbb{R}^{h}$ are the hidden state for $w_i$ and $w_j$. Also, $\bm{v}\in \mathbb{R}^{d}$, $\bm{U} \in \mathbb{R}^{d \times h}$, and $\bm{W} \in \mathbb{R}^{d \times h}$ are trainable parameters.
The probability that $\langle w_j, w_i\rangle$ has the label $l$ is then calculated by
\begin{equation}
\begin{split}
&P_{label}(l|\langle w_j, w_i\rangle, S)\\
&=\frac{\exp(g(l,\langle w_j, w_i\rangle))}{\sum_{k\in \mathcal{L}_{dp}} \exp(g(k,\langle w_j, w_i\rangle))}, 
\end{split}
\end{equation}
where $\mathcal{L}_{dp}$ is the set of labels.
The score function $g$ is expressed as follows.
\begin{equation}
g(l,\langle w_j, w_i\rangle) = \bm{u}_l^T \\ \tanh(\bm{U} \bm{x}_j + \bm{W} \bm{x}_i),
\end{equation}
where $\bm{u}_l\in \mathbb{R}^{d}$ are trainable parameters. \\
We use cross-entropy loss to train this model. The loss is
\begin{equation}
\begin{split}
& J_{dp}(\theta_{dp}) = \\
&\frac{-1}{\sum_{S\in \mathcal{B}}N_S}\sum_{S \in \mathcal{B}} \sum_{i=1}^{N_S} (\log{P(h(w_i)|w_i,S)}\\ 
&+\log{P(l(\langle h(w_i),w_i\rangle)|\langle h(w_i),w_i\rangle,S)}),
\end{split}
\end{equation}
where $\mathcal{B}$ represents one batch, $N_S$ is the number of morphemes in the sentence $S$, $h(w_i)$ is the true head of $w_i$, $l(\langle h(w_i),w_i\rangle)$ is the true label of $\langle h(w_i),w_i\rangle$, and $\theta_{dp}$ are the parameters of the shared encoder (BERT) and the DP decoder. When testing, we first predict the head of $w_i$ as 
\begin{equation}
\hat{h}(w_i) = \argmax_{w_j}(P_{head}(w_j|w_i, S))
\end{equation}
and then predict the label of the $\langle \hat{h}(w_i), w_i\rangle$ edge as
\begin{equation}
\begin{split}
&\hat{l}(\langle \hat{h}(w_i), w_i\rangle) = \\ &\argmax_{l}(P_{label}(l|\langle \hat{h}(w_i),w_i\rangle,S)).
\end{split}
\end{equation}

\subsection{Semantic Role Labeling}
Semantic Role Labeling (SRL) can be broken down into three subtasks: predicate detection, argument identification, and argument classification.
In this paper, the predicate is assumed to be already given, so we only discuss argument identification and argument classification.

The outputs of BERT for the sentence ($X_{srl} = \{\bm{x}_{[CLS]}, \bm{x}_1, ..., \bm{x}_n, \bm{x}_{[SEP]}\}$) are combined with the predicate indicators, which are the features that indicate whether $\bm{x}_i$ is a predicate or not, and the combined results are passed to a one-layer BiLSTM to obtain hidden states $G = \{\bm{g}_{[CLS]}, \bm{g}_1, ..., \bm{g}_n, \bm{g}_{[SEP]}\}$ as
\begin{equation}
    G = BiLSTM(X_{srl}).
\end{equation}

\begin{CJK}{UTF8}{min}
$G$ is transformed into a suitable unit $G' = \{\bm{g'}_1, \bm{g'}_2, ..., \bm{g'}_{n'}\}$ (LUW in the \textit{morpheme setting} and argument span in the \textit{span given setting}).
For the final prediction on each token $\bm{g'}_i$, the token of predicate $\bm{g'}_{p'}$ is concatenated to $\bm{g'}_i$, and then fed into a one-hidden-layer MLP classifier over the label set. Then we obtain $H = \{\bm{h}_1, \bm{h}_2, ..., \bm{h}_{n'}\}$ as
\begin{equation}
    \bm{h}_i = MLP(\bm{g'}_i \oplus \bm{g'}_{p'}),
\end{equation}
where $\bm{h}_i \in \mathbb{R}^{|\mathcal{L}_{srl}|}$, $\oplus$ represents concatenation, and $\mathcal{L}_{srl}$ is the set of labels.
We apply a softmax function to obtain the probability per label as
\begin{equation}
    P_{srl}(l|w_{i}',S') = \frac{\exp(\bm{h}_{i}[l])}{\sum_{k\in \mathcal{L}_{srl}} \exp(\bm{h}_{i}[k])},
\end{equation}
where $S'= \{w_1', ..., w_{n'}'\}$ is the  tokenized sentence for appropriate unit when performing SRL, and $
\bm{h}_{i}[l]$ is the the value corresponding to the label $l$ in the vector $\bm{h}_{i}$.
For example, in Figure \ref{fig:ud-model}, in order to estimate the label of "方" (that person) , the hidden vector of "方" (that person)  is concatenated to the hidden vector of the predicate "切り替え" (switch). Then, this vector is fed into the MLP classifier, and $P_{srl}(l|$方$,S')$ is calculated for each label $l$.
\end{CJK}

Letting the correct label for each morpheme $w_{i}'$ be $r(w_{i}')$, the cross entropy loss is defined as
\begin{equation}
    J_{srl}(\theta_{srl}) =\frac{-1}{N_{\mathcal{B}}} \sum_{S' \in \mathcal{B}}\sum_{w_{i}' \in S'}\log P_{srl}(r(w_{i}')|w_{i}',S'),
\end{equation}
where $\mathcal{B}$ represents one batch, and $N_{\mathcal{B}}$ is the number of sentences in the batch $\mathcal{B}$.
$\theta_{srl}$ are the parameters of the shared encoder (BERT), Bi-LSTM encoder, and SRL decoder.
The prediction of the $r(w_{i}')$ is
\begin{equation}
    \hat{r}(w_{i}') = \argmax_{r}P_{srl}(r | w_i', S')
\end{equation}.

\section{Detailed Setup}

\subsection{SUW and LUW}
There is no unified dictionary of morphemes nor morphological analyzer.
For example, the nodes of UD graph in Universal Dependencies (UD) Japanese treebank are UniDic Short Unit Word (SUW) morphemes\footnote{https://unidic.ninjal.ac.jp/}, BCCWJ-PT adopts Unidic Long Unit Word (LUW) as the morphological unit, and NICT BERT Japanese Pre-trained Model\footnote{https://alaginrc.nict.go.jp/nict-bert/index.html} (NICT BERT) assumes the input is tokenized by MeCab for Juman morphemes.

\begin{CJK}{UTF8}{min}
\begin{figure}
    \centering
    \includegraphics[clip,width=7.5cm]{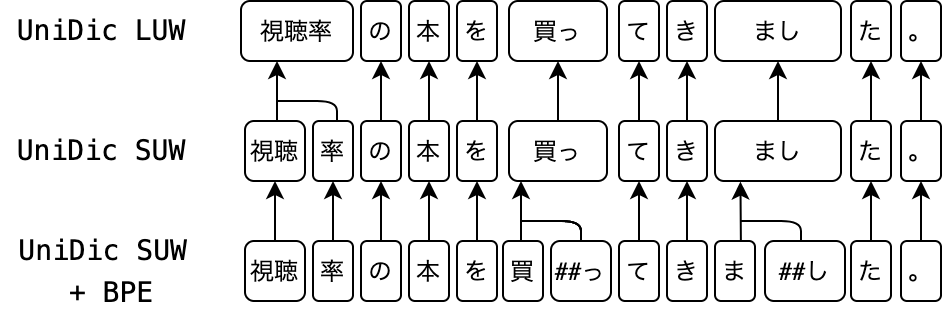}
  \caption{An example of morphological units. The input sentence is "視聴率の本を買ってきました。" (I have bought a book about audience rate.).}
  \label{fig:relation-morphemes}
\end{figure}
Figure \ref{fig:relation-morphemes} shows the example of averaging vectors. The hidden vector of "視聴率" (audience rating) is calculated as the average of the vectors of "視聴" (audience) and "率" (rating).
\end{CJK}

\subsection{Training}
During hyperparameter optimization, we run 3 epochs with the UD train dataset for the DP and DP ROOT models and 10 epochs with the SRL train dataset for others. On the other hand, during full training, we run 10 epochs with the UD train dataset for the DP and DP ROOT models and 30 epochs with the SRL train dataset for others and take test score of each model when the valid score reach the highest. We choose UAS as the valid score for the \textbf{DP} and \textbf{DP known} models, and \textbf{micro $F1$} score for the other models.
To reduce the optimization time, we carry out pruning if the specified validation score is not reached within a certain number of epochs. 
All the time, we set the batch size to be 32, use the AdamW\citep{loshchilov2017decoupled} optimizer with learning rate $\eta$ and carry out the warmup in the first epoch. The maximum number of input tokens is 270 during hyperparameter optimization and 320 during full training.

\section{Detailed Scores}
\begin{figure*}
\centering
  \includegraphics[width=14cm]{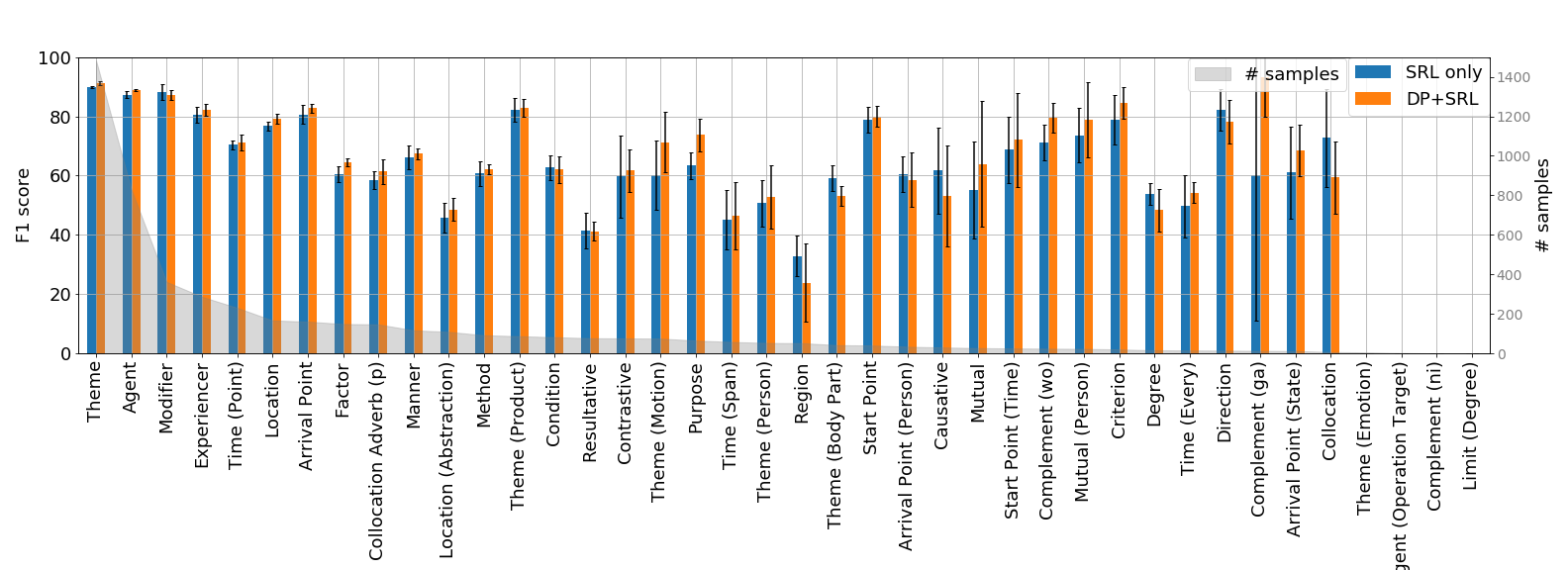}
  \caption{The $F1$ score of each label in the \textit{morpheme setting}. \# samples is number of samples at training phase in each label.}
  \label{fig:f1-per-label}
\end{figure*}

Table\ref{tab:detailed_dp} shows the detailed results for Dependency Parsing(DP) including ablation study settings. Table\ref{tab:detailed_span} shows the detailed results for Semantic Role Labeling(SRL) in the \textit{span given setting}. Table\ref{tab:detailed_mor} shows the detailed results for SRL in the \textit{morpheme setting} including ablation study settings. We also calculated the scores for each label and counted how many labels the multitask model performed better than the stand-alone model and vice versa. The results are shown in Table \ref{tab:label_mor_count}. For these tables the standard deviations of 5 trials are also shown.
Figure \ref{fig:f1-per-label} shows the F1 scores of each semantic role for the \textit{morpheme setting}.

\begin{table}[]
    \centering
    \begin{tabular}{l||ll}
        setting & SRL & DP+SRL \\
        \hline \hline
        morpheme & $11$ & $25$ \\
        span given & $10$ & $26$ 
    \end{tabular}
    \caption{Number of labels of better scores for each type of model}
    \label{tab:label_mor_count}
\end{table}

\begin{table*}
\small
\begin{tabular}{llll}
\hline
\textbf{Model} & \textbf{UAS} & \textbf{LAS} & \textbf{ROOT} \\
\hline
\textbf{DP}  & $\bm{94.84(\pm{0.28})}$ & $93.23(\pm{0.36})$  & $97.10(\pm{0.26})$ \\
\textbf{DP ROOT} & $94.82(\pm{0.11})$ & $\bm{93.29(\pm{0.15})}$   & $\bm{97.23(\pm{0.03})}$ \\
\textbf{SRL$+$DP span given} & $94.54(\pm{0.09})$  & $92.92(\pm{0.12})$  & $\bm{97.23(\pm{0.04})}$  \\
\textbf{SRL$+$DP morpheme} & $89.04(\pm{0.29})$  & $85.94(\pm{0.31})$  & $96.56(\pm{0.20})$ \\
\textbf{- SRL predicate} & $86.03(\pm{0.52})$ & $82.36(\pm{0.58})$ & $95.98(\pm{0.04})$\\
\textbf{- BPE} & $87.49(\pm{1.44})$ & $84.09(\pm{1.80})$ & $94.80(\pm{0.37})$\\
\textbf{+ juman} & $86.88(\pm{1.10})$ & $83.30(\pm{1.37})$ & $96.12(\pm{0.21})$\\
\textbf{- BiLSTM} & $88.14(\pm{2.37})$ & $84.88(\pm{2.68})$ & $96.37(\pm{0.30})$\\
\textbf{- DP predicate} & $84.81(\pm{2.36})$ & $80.99(\pm{2.56})$ & $88.32(\pm{2.26})$\\
\hline
\end{tabular}
\caption{The detailed results for Dependency Parsing.}
\label{tab:detailed_dp}
\end{table*}

\begin{table*}
\small
\begin{tabular}{lllll}
\hline
\textbf{Model} & \textbf{micro $F1_{spa}$(acc.)} & \textbf{macro $F1_{spa}$} & \textbf{macro $Precision$} & \textbf{macro $Recall$}\\
\hline
\textbf{SRL span given} & $75.98(\pm{1.12})$ & $58.74(\pm{1.34})$ & $60.27(\pm{1.05})$ & $60.25(\pm{1.48})$ \\
\textbf{SRL span given(65:5:30)} & $75.19(\pm{0.10})$ & $49.72(\pm{0.68})$ & $50.86(\pm{0.93})$ & $50.72(\pm{0.68})$ \\
\textbf{SRL$+$DP span given} & $\bm{77.48(\pm{0.34})}$ & $\bm{60.68(\pm{0.66})}$ & $\bm{62.15(\pm{1.27})}$ & $\bm{62.10(\pm{1.08})}$ \\
\hline
\end{tabular}
\caption{The detailed results for the \textit{span given setting}.}
\label{tab:detailed_span}
\end{table*}

\begin{table*}
\small
\begin{tabular}{p{1.8cm}llllll}
\hline
\textbf{Model} & \textbf{micro $F1_{mor}$} & \textbf{micro $Precision$} & \textbf{micro $Recall$} & \textbf{macro $F1_{mor}$} & \textbf{macro $Precision$} & \textbf{macro $Recall$}\\
\hline
\textbf{SRL morpheme} & $58.55(\pm{0.54})$ &$55.47(\pm{1.07})$ & $62.00(\pm{0.40})$ & $46.04\pm{0.71})$ & $46.35(\pm{1.45})$ & $49.59(\pm{1.12})$\\
\textbf{SRL$+$DP morpheme} & $60.12(\pm{0.52})$ & $57.22(\pm{0.62})$ & $63.32(\pm{0.51})$ & $47.77(\pm{1.70})$ & $47.89(\pm{2.29})$ & $51.45(\pm{1.39})$ \\
\textbf{- SRL predicate} & $51.49(\pm{0.46})$ & $47.71(\pm{1.00})$ & $55.95(\pm{0.39})$ & $40.71(\pm{2.12})$ & $39.87(\pm{3.11})$ & $44.62(\pm{1.65})$  \\
\textbf{- BPE} & $56.37(\pm{0.61})$ & $53.72(\pm{0.38})$ & $59.29(\pm{0.99})$ & $40.92(\pm{2.25})$ & $40.59(\pm{2.58})$ & $44.09(\pm{2.40})$\\
\textbf{+ juman} & $56.76(\pm{0.96})$ & $53.63(\pm{1.18})$ & $60.28(\pm{0.88})$ & $44.83(\pm{1.72})$ & $45.54(\pm{2.19})$ & $47.96(\pm{1.76})$ \\
\textbf{- BiLSTM} & $59.37(\pm{0.87})$ & $56.53(\pm{1.14})$ & $62.53(\pm{0.99})$ & $46.65(\pm{2.35})$ & $46.20(\pm{1.97})$ & $49.91(\pm{2.84})$ \\
\textbf{- DP predicate} & $59.51(\pm{0.54})$ & $56.97(\pm{0.54})$ & $62.29(\pm{0.60})$ & $46.44(\pm{1.16})$ & $46.30(\pm{1.75})$ & $49.12(\pm{1.17})$ \\
\hline
\end{tabular}
\caption{The detailed results for the \textit{morpheme setting}, including ablation study. Note that in \textbf{- SRL predicate}, we only trained the model for 4 times.}
\label{tab:detailed_mor}
\end{table*}

\section{Ablation study}
\begin{table}[]
    \centering
    \small
    \begin{tabular}{lcc}
        \hline
        Setting & micro $F1_{mor}$ & macro $F1_{mor}$ \\ \hline \hline
        DP+SRL morpheme & $60.12$ & $47.77$ \\
        - SRL predicate & $51.50(-8.62)$ & $39.87(-7.90)$  \\
        - BPE & $56.37(-3.75)$ & $40.93(-6.84)$ \\
        + juman & $56.76(-3.36)$ & $42.89(-4.88)$ \\
        - BiLSTM & $59.38(-0.74)$ & $46.65(-1.12)$ \\
        - DP predicate & $59.51(-0.61)$ & $44.69(-3.08)$ \\
        \hline
    \end{tabular}
    \caption{The results of the ablation study. We trained the models five times with the best hyperparameters for \textbf{DP+SRL morpheme} and averaged the results. }
    \label{tab:ablation_result}
\end{table}
To observe the effects of each individual component of the model, we conducted an ablation study and summarized the results in Table \ref{tab:ablation_result}. In the - BiLSTM setting, we eliminated the BiLSTM encoder and perform DP and SRL at the same layer (hidden layer 1 in Table \ref{fig:ud-model}). In the + juman setting, we tokenized sentences by MeCab with mecab-jumandic for the SRL model instead of using mecab-unidic. - DP predicate and -SRL predicate represent that predicate tokens and a [SEP] token after the first [SEP] token were eliminated from the inputs of DP and SRL respectively (for example, in - SRL predicate, $\{[CLS], w_1, ..., w_n, [SEP]\}$ was used as input for the SRL model instead of $S_{srl}$). In the - BPE setting, the input of tokens was not subword-tokenized by BPE and for that setting we use the NICT BERT model without BPE as the base model. 

As in Table \ref{tab:ablation_result}, SRL predicate has the most crucial effect on the performance of our model. This implies that allowing BERT to incorporate the interaction between the entire sentence and predicate is vital to perform SRL. 
On the other hand, even without the DP predicate, the model's F1 scores did not go down so much. Thus, our model could achieve a high F1 score by using $S_{dp}$ as input on behalf of $S_{multi\_dp}$.
\end{document}